\documentclass[conference]{IEEEtran}
\IEEEoverridecommandlockouts
\usepackage{cite}
\usepackage{amsmath,amssymb,amsfonts}
\usepackage{algorithmic}
\usepackage{graphicx}
\usepackage{textcomp}
\usepackage{xcolor}

\usepackage{colortbl}
\usepackage{tabularx}
\usepackage{listings}

\usepackage{tcolorbox}

\usepackage{booktabs}
\usepackage{pifont}
\usepackage{listings}
\usepackage{caption}
\usepackage{subcaption}
\usepackage{hyperref}
\usepackage[belowskip=-2pt,aboveskip=6pt]{caption}

\definecolor{codegreen}{rgb}{0,0.6,0}
\definecolor{codegray}{rgb}{0.5,0.5,0.5}
\definecolor{codepurple}{rgb}{0.58,0,0.82}
\definecolor{backcolour}{rgb}{0.95,0.95,0.92}

\lstdefinestyle{mystyle}{
    backgroundcolor=\color{backcolour},   
    commentstyle=\color{codegreen},
    keywordstyle=\color{magenta},
    numberstyle=\tiny\color{codegray},
    stringstyle=\color{codepurple},
    basicstyle=\ttfamily\footnotesize,
    breakatwhitespace=false,         
    breaklines=true,                 
    captionpos=b,                    
    keepspaces=true,                 
    numbers=left,                    
    numbersep=5pt,                  
    showspaces=false,                
    showstringspaces=false,
    showtabs=false,                  
    tabsize=2,
    xleftmargin=1.2em 
}

\def\BibTeX{{\rm B\kern-.05em{\sc i\kern-.025em b}\kern-.08em
    T\kern-.1667em\lower.7ex\hbox{E}\kern-.125emX}}
\begin{document}

\title{Advances in \textsc{Appfl}: A Comprehensive and Extensible Federated Learning Framework
}

\author{\IEEEauthorblockN{
Zilinghan Li\IEEEauthorrefmark{1},
Shilan He\IEEEauthorrefmark{2},
Ze Yang\IEEEauthorrefmark{2},
Minseok Ryu\IEEEauthorrefmark{3},
Kibaek Kim\IEEEauthorrefmark{1},
Ravi Madduri\IEEEauthorrefmark{1}
}
\IEEEauthorblockA{
\IEEEauthorrefmark{1}Argonne National Laboratory 
\IEEEauthorrefmark{2}University of Illinois at Urbana-Champaign 
\IEEEauthorrefmark{3}Arizona State University
}
\IEEEauthorblockA{
\{zilinghan.li, kimk, madduri\}@anl.gov, \{shilanh2, zeyang2\}@illinois.edu, minseok.ryu@asu.edu
}
}

\maketitle

\begin{abstract}
Federated learning (FL) is a distributed machine learning paradigm enabling collaborative model training while preserving data privacy. In today's landscape, where most data is proprietary, confidential, and distributed, FL has become a promising approach to leverage such data effectively, particularly in sensitive domains such as medicine and the electric grid. Heterogeneity and security are the key challenges in FL, however, most existing FL frameworks either fail to address these challenges adequately or lack the flexibility to incorporate new solutions. To this end, we present the recent advances in developing \textsc{Appfl}, an extensible framework and benchmarking suite for federated learning, which offers comprehensive solutions for heterogeneity and security concerns, as well as user-friendly interfaces for integrating new algorithms or adapting to new applications. We demonstrate the capabilities of \textsc{Appfl} through extensive experiments evaluating various aspects of FL, including communication efficiency, privacy preservation, computational performance, and resource utilization. We further highlight the extensibility of \textsc{Appfl} through case studies in vertical, hierarchical, and decentralized FL. \textsc{Appfl} is fully open-sourced on GitHub at \href{https://github.com/APPFL/APPFL}{https://github.com/APPFL/APPFL}.
\end{abstract}


\section{Introduction}
Availability of extensive training data is becoming increasingly crucial for developing more capable machine learning (ML) models, especially as these models continue to grow in size and complexity. Nonetheless, most of the data in today's landscape is confidential and distributed across various data silos \cite{crawford2021atlas}. This distribution makes it difficult to collect the data for centralized model training, posing significant challenges in fully leveraging the existing data to train more powerful ML models. In this context, federated learning (FL), a distributed ML paradigm, offers a promising solution to utilize data from multiple data owners without direct data sharing \cite{fedavg, FLadvances}. 

In FL, multiple data owners, referred to as clients, collaborate under a central server to train a shared ML model by iterating two steps: (1) each client trains an ML model using its local dataset and submits the updated model to the server, and (2) the server aggregates these local models to update the global model and then sends it back to the clients for further local training. In this way, FL leverages data from multiple sources to build a more powerful and robust model without data centralization, thereby protecting data privacy. FL has been widely adopted in domains such as medicine \cite{fl_cancer, hoang2023enabling}, finance \cite{fl_insurance}, and electric grid \cite{bose2023federated}, where data privacy is paramount. Depending on the amount, capability, and availability of client devices, FL is broadly categorized into two types, cross-device FL and cross-silo FL \cite{FLadvances}. In cross-device FL, numerous mobile or IoT devices with limited computing power and intermittent availability collaboratively train relatively small models such as keyboard suggestion models \cite{hard2018federated}. In contrast, cross-silo FL involves fewer but more reliable and powerful clients, typically represented by large data silos and institutions, to develop more complex ML models with extensive parameters. 

While FL can be conceptually simplified to traditional machine learning with an additional global aggregation operation, its distributed nature introduces significant challenges in terms of heterogeneity and security. Data heterogeneity, stemming from the unbalanced, and non-independent and identically distributed (non-IID) nature of client local datasets, can lead to varied local training objectives across clients and potentially degrade the performance of the global model \cite{fedavgm}. Additionally, the heterogeneity in computation and communication, caused by diverse computing capabilities and network connectivity of client devices, can severely impact the efficiency of FL training \cite{xie2019asynchronous}. This is particularly problematic in synchronous FL algorithms, where the server has to wait for all clients to submit their local models before global aggregation. With regard to security, FL is vulnerable to various attacks. Untrusted clients might maliciously attack FL experiments by submitting corrupted local models, and there is also a risk that training data could be reconstructed from the model updates sent by clients, thereby compromising data privacy \cite{FLadvances, kaissis2021end}. 

Most existing FL frameworks, such as \textsc{Flower} \cite{beutel2020flower}, \textsc{FedML} \cite{he2020fedml}, and \textsc{FedScale} \cite{lai2022fedscale}, do not adequately address the full spectrum of FL challenges. For example, some frameworks do not support asynchronous aggregation that could improve training efficiency, lack implementations of robust authentication, or fail to offer user-friendly interfaces for easy integration of new algorithms.
To bridge these gaps, we developed the Advanced Privacy-Preserving Federated Learning (\textsc{Appfl}) framework, a comprehensive and extensible FL framework that builds on and improves the work presented in \cite{ryu2022appfl}. 
\textsc{Appfl} features advanced aggregation strategies to address data heterogeneity \cite{fedavgm} and various asynchronous aggregation strategies to boost training efficiency among heterogeneous computing resources \cite{li2023fedcompass}. Additionally, \textsc{Appfl} incorporates versatile communication protocols, data transfer methods, and compression strategies to meet different communication requirements and enhance communication efficiency. It also includes robust authentication via Globus \cite{tuecke2016globus}, along with plugins for adding new authentication methods, and implements privacy preservation strategies \cite{dwork2006differential} to prevent the reconstruction of training data. Moreover, \textsc{Appfl} is extensible; it follows a modular design with detailed documentation  that enables users and developers to seamlessly adapt the framework for different application use cases and integrate custom algorithmic solutions to tackle various FL challenges.

The contributions of this work are outlined as follows:
\begin{itemize}
    \item Advance \textsc{Appfl}, an open-source and well-documented\footnote{The documentation of \textsc{Appfl} is available at  \href{https://appfl.ai}{https://appfl.ai}.} FL framework for both FL users and developers that provides established solutions to common FL challenges for FL users and offers flexible and modular interfaces facilitating easy integration of new algorithmic solutions for FL developers 
    \item Conduct comprehensive evaluations of various aspects of FL using \textsc{Appfl}, including the efficiency of the versatile communication protocols, data transfer methods, and compression strategies, as well as the performance of privacy preservation strategies and training effectiveness of different FL aggregation algorithms 
    \item Provide case studies in vertical, hierarchical, decentralized FL to highlight the extensibility and adaptability of the \textsc{Appfl} framework in diverse FL scenarios
\end{itemize}

\section{Background and Related Work}
\subsection{Heterogeneity in Federated Learning}
Heterogeneity is one of the key challenges in FL due to its distributed nature. This heterogeneity can be categorized into three primary types: data heterogeneity, computation heterogeneity, and communication heterogeneity.

\textit{Data heterogeneity} arises from the fact that client datasets are unbalanced and non-IID, meaning they may not be representative of the overall population. This discrepancy leads to varying local training objectives among clients, causing their locally trained models to diverge from one another, a phenomenon known as client drift \cite{karimireddy2020scaffold}. As a result, simple weighted averaging of local models, as in the \texttt{FedAvg} strategy \cite{fedavg}, may degrade the performance of the global model as data heterogeneity increases \cite{fedavgm}. Several solutions have been proposed to address this issue on both the server and client sides. For instance, server-side optimizations such as \texttt{FedAvgM} \cite{fedavgm}, \texttt{FedAdam}, \texttt{FedAdagrad}, and \texttt{FedYogi} \cite{reddi2020adaptive} have been introduced to enhance FL performance on non-IID data. Client-side approaches such as \texttt{SCAFFOLD} \cite{karimireddy2020scaffold} incorporate correction terms into the client's local objective function to reduce drift between local and global models. Additionally, selecting participating clients based on data quality and relevance has been explored as well as an effective solution \cite{hiniduma2024ai, cao2022birds, pan2023contextual}.

\textit{Computation heterogeneity} occurs when the computing devices of FL clients have varying computing power, resulting in large variants in the local training times. This variance poses challenges for synchronous FL strategies, where the server must wait for all clients to submit their local models before global aggregation. Delays from slower clients can reduce the overall training efficiency and lead to underutilization of computing resources. In order to address this issue, various asynchronous aggregation strategies have been proposed. These strategies, including \texttt{FedAsync} \cite{xie2019asynchronous}, \texttt{FedBuff} \cite{nguyen2022federated}, and \texttt{FedCompass} \cite{li2023fedcompass}, update the global model immediately upon receiving models from one or a few clients. These methods are beneficial in environments with heterogeneous computing capabilities as they minimize client idle time. Other approaches include disregarding contributions from straggler clients \cite{bonawitz2019towards} or explicitly selecting clients for local training based on their computing capabilities \cite{ chen2021towards}. Nonetheless, these methods are best suited for cross-device FL, where only a subset of clients participates in each training round. They do not align well with cross-silo FL where there are only a few FL clients and ensuring the participation of every client is vital for maintaining the robustness of the global model.

\textit{Communication heterogeneity} is originally rooted in the intermittent availability of client devices due to the limited power and bandwidth in cross-device FL, which is less of an issue in cross-silo FL. However, as foundation models increasingly dominate various domains, the interest in using FL to train or fine-tune these models has surged. This surge has led to a substantial increase in communication costs, which become a critical factor affecting the FL training efficiency \cite{li2024secure}. Consequently, improving the efficiency and robustness of transferring large model parameters has become critically important as well\cite{chen2023federated}. To address this situation, some client selection methods have been proposed to mitigate communication issues in cross-device FL by strategically selecting clients based on their availability, data quality, and performance \cite{cho2020client}. Other approaches focus on generic FL settings by applying compression or pruning techniques to large model parameters, thereby reducing the communication workload \cite{bai2024fedspallm, wilkins2024fedsz}.

\vspace{-3pt}

\subsection{Attacks and Security Concerns in Federated Learning}
\begin{table*}[htbp]
\captionsetup{belowskip=-17pt} 
\caption{Comparison of popular open-source federated learning frameworks.}
\label{tab:frameworks}
\centering
\setlength{\tabcolsep}{2.7pt}
\resizebox{\linewidth}{!}{\begin{tabular}{r c c c c c c c c c c} 
\toprule 
Framework & Data Hetero. & Sync. FL & Async. FL & Compression & Versatile Comm. & Privacy & Auth. & Real Deployment & FL Variants \\
\midrule 
\textsc{LEAF} \cite{caldas2018leaf}& \textcolor[rgb]{0.9, 0.3, 0.3}{\ding{55}} & \textcolor[rgb]{0.3, 0.7, 0.3}{\ding{51}} & \textcolor[rgb]{0.9, 0.3, 0.3}{\ding{55}} & \textcolor[rgb]{0.9, 0.3, 0.3}{\ding{55}} & \textcolor[rgb]{0.9, 0.3, 0.3}{\ding{55}} & \textcolor[rgb]{0.9, 0.3, 0.3}{\ding{55}} & \textcolor[rgb]{0.9, 0.3, 0.3}{\ding{55}} & \textcolor[rgb]{0.9, 0.3, 0.3}{\ding{55}} & \textcolor[rgb]{0.9, 0.3, 0.3}{\ding{55}}\\
\textsc{TFF} \cite{bonawitz2019towards} & \textcolor[rgb]{0.3, 0.7, 0.3}{\ding{51}} & \textcolor[rgb]{0.3, 0.7, 0.3}{\ding{51}} & \textcolor[rgb]{0.9, 0.3, 0.3}{\ding{55}} & \textcolor[rgb]{0.9, 0.3, 0.3}{\ding{55}} & \textcolor[rgb]{0.9, 0.3, 0.3}{\ding{55}} & \textcolor[rgb]{0.3, 0.7, 0.3}{\ding{51}} & \textcolor[rgb]{0.9, 0.3, 0.3}{\ding{55}} & \textcolor[rgb]{0.9, 0.3, 0.3}{\ding{55}} & \textcolor[rgb]{0.9, 0.3, 0.3}{\ding{55}}\\
\textsc{Appfl-v0} \cite{ryu2022appfl} & \textcolor[rgb]{0.3, 0.7, 0.3}{\ding{51}} & \textcolor[rgb]{0.3, 0.7, 0.3}{\ding{51}} & \textcolor[rgb]{0.9, 0.3, 0.3}{\ding{55}} & \textcolor[rgb]{0.9, 0.3, 0.3}{\ding{55}} & \textcolor[rgb]{0.9, 0.3, 0.3}{\ding{55}} & \textcolor[rgb]{0.3, 0.7, 0.3}{\ding{51}} & \textcolor[rgb]{0.9, 0.3, 0.3}{\ding{55}} & \textcolor[rgb]{0.3, 0.7, 0.3}{\ding{51}} & \textcolor[rgb]{0.9, 0.3, 0.3}{\ding{55}}\\
\textsc{FederatedScope} \cite{xie2022federatedscope} & \textcolor[rgb]{0.3, 0.7, 0.3}{\ding{51}} & \textcolor[rgb]{0.3, 0.7, 0.3}{\ding{51}} & \textcolor[rgb]{0.9, 0.3, 0.3}{\ding{55}} & \textcolor[rgb]{0.9, 0.3, 0.3}{\ding{55}} & \textcolor[rgb]{0.9, 0.3, 0.3}{\ding{55}} & \textcolor[rgb]{0.3, 0.7, 0.3}{\ding{51}} & \textcolor[rgb]{0.9, 0.3, 0.3}{\ding{55}} & \textcolor[rgb]{0.3, 0.7, 0.3}{\ding{51}}& \textcolor[rgb]{0.3, 0.7, 0.3}{\textbf{\textit{VFL}}} \\
\textsc{FLARE} \cite{roth2022nvidia} & \textcolor[rgb]{0.3, 0.7, 0.3}{\ding{51}} & \textcolor[rgb]{0.3, 0.7, 0.3}{\ding{51}} & \textcolor[rgb]{0.9, 0.3, 0.3}{\ding{55}} & \textcolor[rgb]{0.9, 0.3, 0.3}{\ding{55}} & \textcolor[rgb]{0.9, 0.3, 0.3}{\ding{55}} & \textcolor[rgb]{0.3, 0.7, 0.3}{\ding{51}} & \textcolor[rgb]{0.3, 0.7, 0.3}{\ding{51}} & \textcolor[rgb]{0.3, 0.7, 0.3}{\ding{51}} & \textcolor[rgb]{0.3, 0.7, 0.3}{\textbf{\textit{VFL}}}\\ 
\textsc{OpenFL} \cite{foley2022openfl} & \textcolor[rgb]{0.3, 0.7, 0.3}{\ding{51}} & \textcolor[rgb]{0.3, 0.7, 0.3}{\ding{51}} & \textcolor[rgb]{0.9, 0.3, 0.3}{\ding{55}} & \textcolor[rgb]{0.3, 0.7, 0.3}{\ding{51}} & \textcolor[rgb]{0.9, 0.3, 0.3}{\ding{55}}  & \textcolor[rgb]{0.3, 0.7, 0.3}{\ding{51}} & \textcolor[rgb]{0.3, 0.7, 0.3}{\ding{51}} & \textcolor[rgb]{0.3, 0.7, 0.3}{\ding{51}} & \textcolor[rgb]{0.3, 0.7, 0.3}{\textbf{\textit{VFL}}}\\ 
\textsc{FedScale} \cite{lai2022fedscale} & \textcolor[rgb]{0.3, 0.7, 0.3}{\ding{51}} & \textcolor[rgb]{0.3, 0.7, 0.3}{\ding{51}} & \textcolor[rgb]{0.3, 0.7, 0.3}{\ding{51}} & \textcolor[rgb]{0.3, 0.7, 0.3}{\ding{51}} & \textcolor[rgb]{0.9, 0.3, 0.3}{\ding{55}} & \textcolor[rgb]{0.3, 0.7, 0.3}{\ding{51}} & \textcolor[rgb]{0.9, 0.3, 0.3}{\ding{55}} & \textcolor[rgb]{0.3, 0.7, 0.3}{\ding{51}} & \textcolor[rgb]{0.9, 0.3, 0.3}{\ding{55}}\\ 
\textsc{FedLab} \cite{zeng2023fedlab} & \textcolor[rgb]{0.3, 0.7, 0.3}{\ding{51}} & \textcolor[rgb]{0.3, 0.7, 0.3}{\ding{51}} & \textcolor[rgb]{0.3, 0.7, 0.3}{\ding{51}} & \textcolor[rgb]{0.3, 0.7, 0.3}{\ding{51}} & \textcolor[rgb]{0.9, 0.3, 0.3}{\ding{55}} & \textcolor[rgb]{0.9, 0.3, 0.3}{\ding{55}} & \textcolor[rgb]{0.9, 0.3, 0.3}{\ding{55}} & \textcolor[rgb]{0.3, 0.7, 0.3}{\ding{51}} & \textcolor[rgb]{0.9, 0.3, 0.3}{\ding{55}}\\ 
\textsc{Flower} \cite{beutel2020flower} & \textcolor[rgb]{0.3, 0.7, 0.3}{\ding{51}} & \textcolor[rgb]{0.3, 0.7, 0.3}{\ding{51}} & \textcolor[rgb]{0.9, 0.3, 0.3}{\ding{55}} & \textcolor[rgb]{0.9, 0.3, 0.3}{\ding{55}} & \textcolor[rgb]{0.3, 0.7, 0.3}{\ding{51}} & \textcolor[rgb]{0.3, 0.7, 0.3}{\ding{51}} & \textcolor[rgb]{0.3, 0.7, 0.3}{\ding{51}} & \textcolor[rgb]{0.3, 0.7, 0.3}{\ding{51}} & \textcolor[rgb]{0.3, 0.7, 0.3}{\textbf{\textit{VFL}}}\\
\textsc{FedML} \cite{he2020fedml} & \textcolor[rgb]{0.3, 0.7, 0.3}{\ding{51}} & \textcolor[rgb]{0.3, 0.7, 0.3}{\ding{51}} & \textcolor[rgb]{0.9, 0.3, 0.3}{\ding{55}} & \textcolor[rgb]{0.9, 0.3, 0.3}{\ding{55}} & \textcolor[rgb]{0.3, 0.7, 0.3}{\ding{51}} & \textcolor[rgb]{0.3, 0.7, 0.3}{\ding{51}} & \textcolor[rgb]{0.3, 0.7, 0.3}{\ding{51}} & \textcolor[rgb]{0.3, 0.7, 0.3}{\ding{51}} & \textcolor[rgb]{0.3, 0.7, 0.3}{\textbf{\textit{VFL, HierFL, DFL}}}\\
\textsc{Appfl} (this work) & \textcolor[rgb]{0.3, 0.7, 0.3}{\ding{51}} & \textcolor[rgb]{0.3, 0.7, 0.3}{\ding{51}} & \textcolor[rgb]{0.3, 0.7, 0.3}{\ding{51}} & \textcolor[rgb]{0.3, 0.7, 0.3}{\ding{51}} & \textcolor[rgb]{0.3, 0.7, 0.3}{\ding{51}} & \textcolor[rgb]{0.3, 0.7, 0.3}{\ding{51}} & \textcolor[rgb]{0.3, 0.7, 0.3}{\ding{51}} & \textcolor[rgb]{0.3, 0.7, 0.3}{\ding{51}} & \textcolor[rgb]{0.3, 0.7, 0.3}{\textbf{\textit{VFL, HierFL, DFL}}}\\
\bottomrule
\vspace{-20pt}
\end{tabular}}
\end{table*}

The distributed and uninspectable nature of FL exposes it to various adversarial attacks and security risks. These attacks generally fall into two broad categories: (1) inferring clients' confidential training data from the model gradients and (2) degrading the performance of the trained global model \cite{FLadvances}. 

Gradient inversion algorithms, for example, can reveal information about the private training data by iteratively updating a randomly initialized sample to match its gradient update with the actual model gradient. These algorithms are particularly effective in the early stages of training where the gradients contain more information about the training data \cite{hatamizadeh2023gradient}. Countermeasures to these inversion attacks include increasing training batch sizes \cite{yin2021see}, implementing differential privacy techniques to add noise to model gradients \cite{hoang2023enabling, kaissis2021end}, and compressing model gradients \cite{ tsuzuku2018variance}.

Since the FL server cannot inspect the client training data, FL is also vulnerable to attacks from Byzantine clients, which may either submit corrupted model parameters (model poisoning) or use tampered data for training (data poisoning) \cite{tolpegin2020data}. To counter these, several algorithms have been developed to exclude models whose parameters significantly deviate from the norm \cite{ cao2020fltrust}. 
Alternatively, some solutions assume that the FL server holds a clean and secret validation dataset to evaluate and potentially exclude poorly performing client models from the aggregation process  \cite{xie2019zeno}. Beyond algorithmic defenses, addressing malicious attacks in FL can also be achieved through system-level enhancements, particularly by integrating with identity and access management (IAM) services \cite{li2023appflx}. Such integration enables the creation of secure federations that permit only trusted and known clients to participate in FL experiments, thus alleviating security concerns at their root.

\vspace{-4pt}

\subsection{Existing Federated Learning Frameworks}
We conduct a brief survey of several popular open-source federated learning frameworks, including the work we built upon without advancements \cite{ryu2022appfl} (denoted as \textsc{Appfl-v0}), focusing on their solutions to heterogeneity and security challenges, usability, and extensibility for various application scenarios. The results are summarized in Table~\ref{tab:frameworks}. 

In addressing data heterogeneity, most frameworks implement advanced client training and server aggregation strategies to mitigate client drift issues, with the exception of \textsc{LEAF} \cite{caldas2018leaf}. Regarding computation heterogeneity, while all frameworks include synchronous FL algorithms, only \textsc{FedScale} \cite{lai2022fedscale}, \textsc{FedLab} \cite{zeng2023fedlab}, and \textsc{Appfl} offer asynchronous communication stack and corresponding asynchronous aggregation strategies. For communication heterogeneity concerns, we evaluate whether the frameworks feature lossless or lossy compression algorithms to reduce the communication loads and whether they provide versatile communication stacks that support multiple protocols, enhancing efficiency and adaptability to different deployment requirements and scenarios.

For the privacy and security challenges, our investigation focuses on whether the frameworks incorporate privacy preservation or enhancing algorithms, and integrate IAM services for user authentication and authorization. Most existing frameworks support privacy preservation to some extent, with \textsc{FLARE} \cite{roth2022nvidia}, \textsc{OpenFL} \cite{foley2022openfl}, \textsc{Flower} \cite{beutel2020flower}, \textsc{FedML} \cite{he2020fedml}, and \textsc{Appfl} featuring IAM integration for verifying user identities and managing access to specific FL experiments.

As for usability, most of the frameworks facilitate both the simulation of FL experiments within the same machine or cluster and the real deployment among distributed clients. However, \textsc{LEAF} and \textsc{TFF} \cite{bonawitz2019towards} are limited to simulation environments only. To evaluate the extensibility and ease of customization of the frameworks, we access their support for different FL variants beyond the traditional federated learning, specifically vertical federated learning (VFL) \cite{liu2024vertical}, hierarchical federated learning (HierFL) \cite{abad2020hierarchical}, and decentralized federated learning (DFL) \cite{lalitha2018fully}. \textsc{FederatedScope} \cite{xie2022federatedscope}, \textsc{FLARE}, \textsc{OpenFL}, and \textsc{Flower} provide use cases in VFL settings, and \textsc{FedML} and \textsc{Appfl} extend support to all three variants. 
\vspace{-4pt}
\section{Framework Architecture and Implementation}
\vspace{-4pt}
\begin{figure}[h]
\centerline{\includegraphics[width=0.98\linewidth]{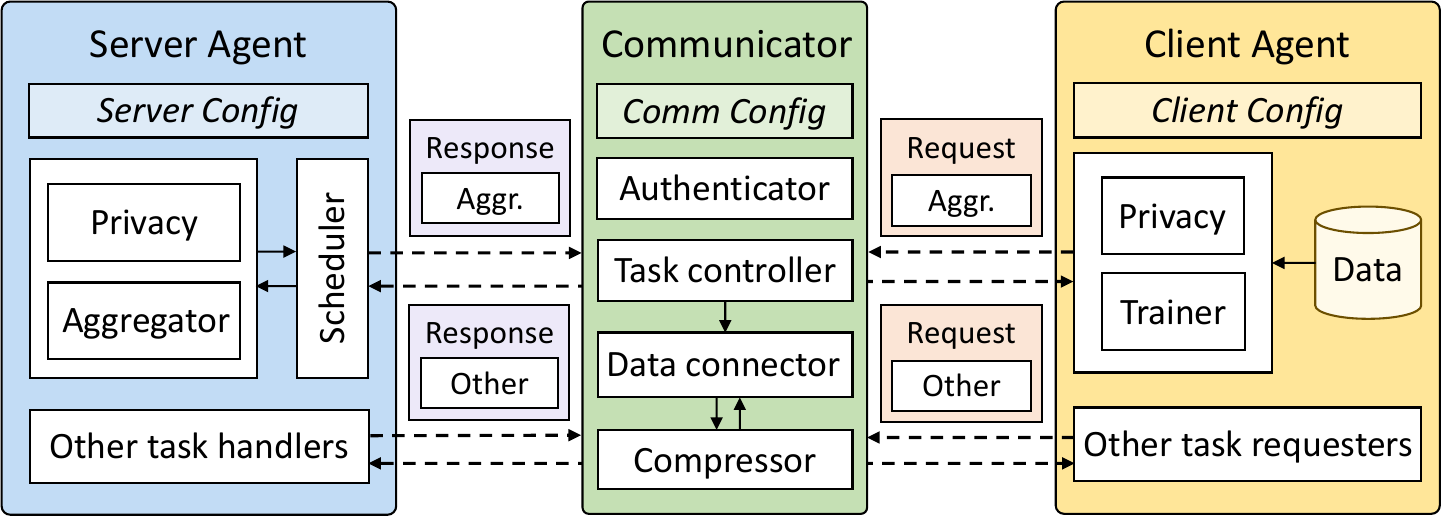}}
\caption{Overview of the \textsc{Appfl} framework's new software architecture design. \textit{Server agent} and \textit{client agent} act on behalf of the FL server and client, respectively, to fulfill various tasks for FL experiments. \textit{Communicator} exchanges task control signals and model parameters between the server and client.}
\label{fig:appfl-overview}
\end{figure}
The \textsc{Appfl} framework is a Python package available on \texttt{PyPI}. Figure~\ref{fig:appfl-overview} provides an overview of its new software architecture. \textsc{Appfl} defines a \textit{server agent} and a \textit{client agent}, connected by the \textit{communicator}, to represent the FL server and clients in performing the primary aggregation task and other necessary tasks for running FL experiments. The \textit{server agent} is mainly composed of a scheduler module that orchestrates the aggregation of client local models under various synchronicity settings, an aggregator module that aggregates the local models passed from the scheduler to update global model, and a privacy module for additional privacy protection. The \textit{client agent} consists of a trainer module responsible for training the ML model using the confidential local dataset and a privacy module for the privacy preservation algorithms. The \textit{communicator} facilitates robust communication between the server and clients, supporting multiple communication protocols for exchanging task control signals and data, with an option to separate the transmission of control signals and data via a data connector. Additionally, the communicator incorporates several compressors for improved efficiency and authenticators for securing the FL experiments. Overall, \textsc{Appfl} incorporates solutions for various challenges in FL and is designed to be modular and extensible, facilitating easy integration of new algorithms and strategies to address FL challenges. The following subsections detail several key components of \textsc{Appfl}.

\subsection{FL Experiment Configuration}
\textsc{Appfl} provides a straightforward way to configure FL experiments: each experiment utilizes a configuration YAML file for the FL server and individual YAML files for each FL client. Listing~\ref{lst:server-config} presents an example of the server configuration file, which includes server-specific settings, such as the aggregation algorithm and the number of global epochs, along with general configurations for the clients like the trainer and compressor types. These general configurations are distributed to all clients at the beginning of each FL experiment, simplifying the setup by ensuring that shared configuration fields do not need to be individually set by each client. 
In addition to the configurations shared by the server, each client possesses its own YAML configuration file that defines client-specific settings, as shown in Listing~\ref{lst:client-config}. The client-specific settings include a Python loader file, which defines a function for loading the client's local datasets, and some training-related configurations such as the device to use and directories for logging and checkpoints. 

This configuration  also facilitates integration of new algorithms by allowing developers to directly add necessary settings to the relevant configuration files and use them in their respective module blocks. For instance, to create a trainer for a particular application, a developer simply needs to define a new trainer within the \textsc{Appfl} trainer module and include all necessary arguments in the \textit{client\_configs.train\_configs} section of the configuration file.

\begin{lstlisting}[caption={An example server configuration YAML file, containing both server configurations and general client configurations to be shared among all clients. }, label={lst:server-config}]
# Server configurations
server_configs:
  aggregator: FedAvgAggregator
  num_global_epochs: 10
  ...
# General client configurations for all clients
client_configs:
  train_configs:
    trainer: VanillaTrainer
    lr: 0.001
    ...
  comm_configs:
    compressor_configs:
        lossy_compressor: SZ2Compressor
        ...
\end{lstlisting}

\begin{lstlisting}[caption={An example client configuration YAML file, containing client-specific configurations such as the data loader file.}, label={lst:client-config}]
# Information needed to load local data
data_configs:
  dataset_path: ./dataset/covid_dataset.py
  dataset_name: get_covid #function to load data
  dataset_kwargs: #optional function arguments
    ...
# Client-specific training settings
train_configs:
  device: cpu
  logging_dir: ./appfl_logging
  checkpoint_dir: ./appfl_checkpoint
  ...
\end{lstlisting}

\vspace{-4.5pt}

\subsection{Communication Stack}
\vspace{-1.5pt}
\begin{figure}[h]
\centerline{\includegraphics[width=0.93\linewidth]{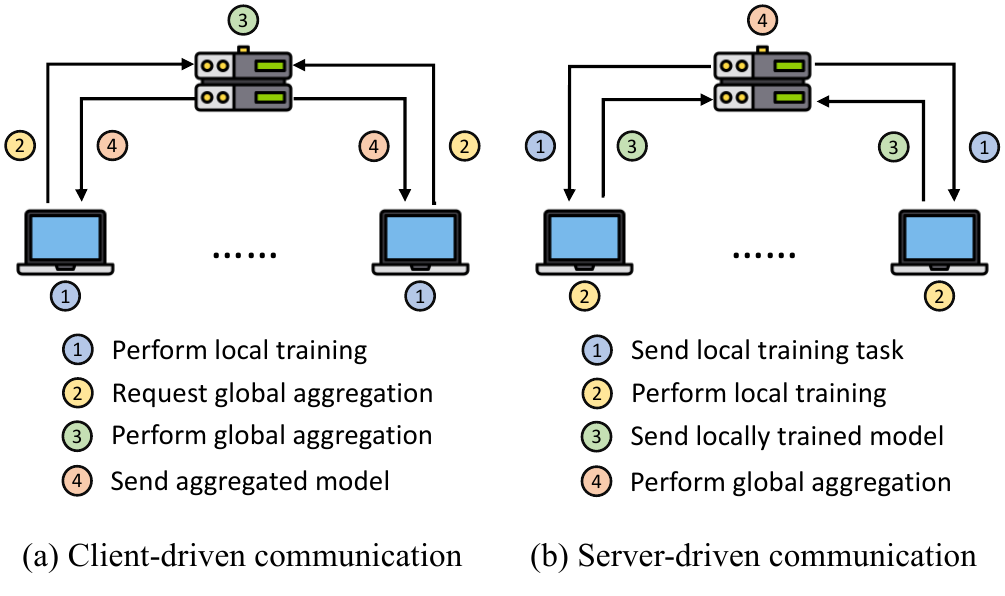}}
\caption{Running one local training and global aggregation iteration using (a) client-driven and (b) server-driven communication protocols.}
\label{fig:communication}
\end{figure}
In FL, communication protocols can be broadly classified into two types based on the driven side of the FL process: (1) \textit{client-driven}: the clients control the FL process and interact with the server for aggregation and other tasks by sending various requests and (2) \textit{server-driven}: the server controls the FL process by dispatching various types of tasks to the clients. Figure~\ref{fig:communication} illustrates the differences between these two types of communication protocols during a local training and global aggregation FL iteration. Client-driven protocols offer clients greater autonomy over the FL process, whereas server-driven protocols simplify the coordination of FL experiments, with the central server itself managing the whole distributed training process. The \textsc{Appfl} communicator supports MPI and gRPC as client-driven communication protocols and Globus Compute \cite{chard2020funcx} as the server-driven protocol. Specifically, MPI is for  simulation purposes only, while gRPC and Globus Compute can be used for real deployments. Notably, gRPC requires the server to open a specific port for inbound TCP connections, which is typically restricted in high-performance computing  environments and institutional computing facilities. Conversely, Globus Compute only necessitates outbound connections to the Globus service, thus enabling a broader range of computing resources to serve as the FL server. The versatile communication protocols supported by \textsc{Appfl} make it capable of meeting diverse communication needs in FL deployments.

For client-driven communication protocols, the \textsc{Appfl} {communicator} provides a server communicator that defines handlers for various types of requests, such as sending configurations and performing global aggregation, by interacting with the server agent. Additionally, a client communicator assists the client agents in sending requests to the server. As for Globus Compute, the server-driven communication protocol, it is a distributed function-as-a-service platform that can dispatch Python functions to run on remote machines. The \textsc{Appfl} communicator provides a Globus Compute server communicator to send various tasks, such as local training, to run on the remote client machines and collect results back for conducting FL experiments. Overall, \textsc{Appfl} supports commonly used server request handlers and client task implementations and provides a user-friendly interface that enables developers to easily define new request handlers or tasks without in-depth knowledge of the underlying communication protocol.

While the communication protocols can transfer the task control signals (i.e., requests in client-driven and tasks in server-driven protocols) along with the associated data, \textsc{Appfl} provides an option to separate the transfer of task controls from the associated model parameters through the integration with ProxyStore \cite{pauloski2023proxystore,pauloski2024proxystore}. ProxyStore can create a \textit{proxy} for any target Python object, providing a lightweight reference that can remotely resolve the target object when used. When the \textit{proxy} is getting resolved, the object is transferred via an underlying data connector from the producer to the consumer. \textsc{Appfl} currently supports two connectors: an S3 connector, which uses AWS S3 buckets for remote data transfer, and a ProxyStore endpoint connector, which transfers data via the ProxyStore-hosted relay server. The integration with ProxyStore offers two main benefits: (1) it prevents exceeding the maximum data size limits imposed by certain communication  protocols (e.g., Globus Compute restricts task arguments and result sizes to 10 MB to reduce its service load, thus making data transfer separation a must when exchanging large model parameters), and (2) it offers users a variety of data transmission options for different communication scenarios and facilitates easy integration of other efficient data transmission methods suitable for their specific use cases to accelerate the FL communication, regardless of the communication protocol in use. 

Furthermore, \textsc{Appfl} incorporates a range of data compressors to enhance communication efficiency, crucial for transferring parameters of large models or operating in environments with limited network bandwidth. It supports various lossless compressors including zstd, gzip, and blosc, as well as lossy data compressors including SZ2 \cite{liang2018error}, SZ3 \cite{liang2022sz3}, and ZFP \cite{lindstrom2014fixed}. These compressors can help reduce the communication load, enabling faster data transfer between the server and clients.

\vspace{-3pt}
\subsection{Server Scheduling and Aggregation}

\vspace{-10pt}
\begin{figure}[!htbp]
\centerline{\includegraphics[width=0.97\linewidth]{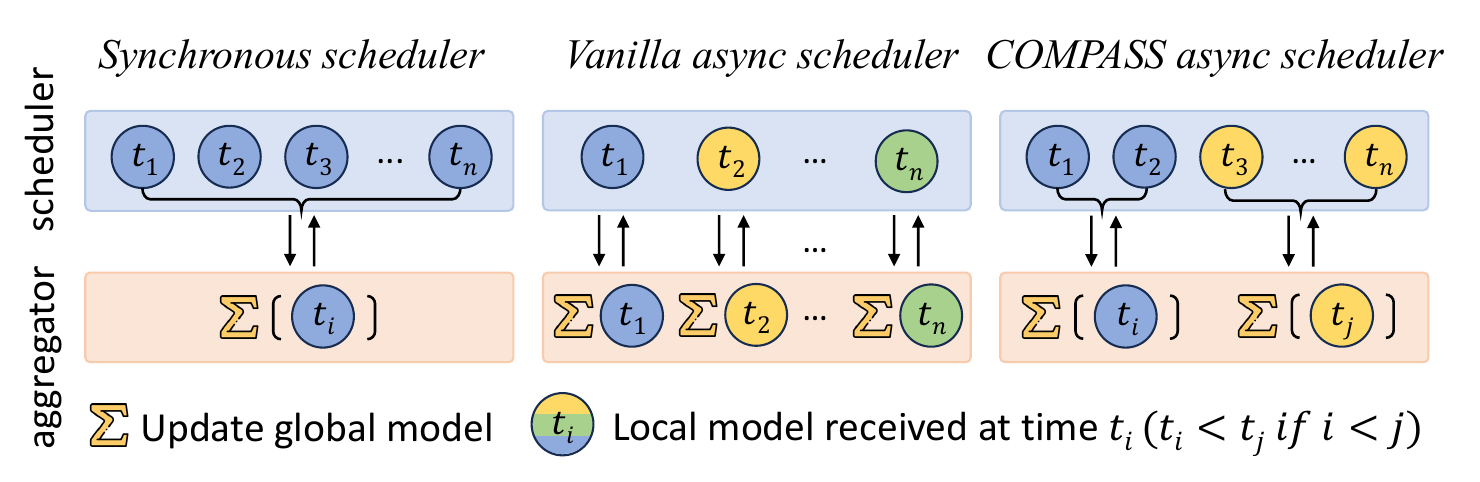}}
    \caption{Scheduling of the aggregation for client local models under three schedulers with different synchronicity settings.}
    \label{fig:scheduler}
\end{figure}
In order to tackle the computation heterogeneity in FL where clients have varying computing capabilities, many asynchronous aggregation algorithms have been proposed to reduce client idle times and enhance resource utilization. To support aggregation with different synchronicity settings, \textsc{Appfl} introduces a server-side scheduler that acts as an interface between the communicator and the aggregator. Upon receiving a local model from a client, the communicator forwards it to the scheduler, which determines the appropriate time to pass the local model(s) to the aggregator for updating the global model. For synchronous aggregation strategies, such as \texttt{FedAvg} \cite{fedavg}, a synchronous scheduler buffers each client's local model until all models are received, at which point it forwards them to the aggregator to update the global model. Conversely, for asynchronous strategies like \texttt{FedAsync} \cite{xie2019asynchronous}, a vanilla asynchronous scheduler immediately sends the client model to the aggregator and returns the updated global model back to the communicator. Additionally, the scheduler module is designed to be extensible for the incorporation of more advanced scheduling algorithms. Specifically, \textsc{Appfl} supports the state-of-the-art \texttt{Compass} asynchronous scheduler \cite{li2023fedcompass}, aiming to alleviate the drift of the global model toward faster clients. Such drift is prevalent in other asynchronous FL algorithms, where faster clients update the global model more frequently and the models from slower clients become stale. The \texttt{Compass} scheduler  synchronizes the arrival of a group of client local models by assigning different amounts of local training tasks to different clients to enable a grouped global aggregation and avoid stale local models, alleviating the client drift issue. The seamless integration of \texttt{FedCompass} further exemplifies our framework’s extensibility. Figure~\ref{fig:scheduler} illustrates the scheduling processes under the three different schedulers. 


As for the aggregator module, \textsc{Appfl} supports a broad range of aggregation strategies, going beyond the widely used \texttt{FedAvg}. These include \texttt{FedAvgM}, \texttt{FedAdam}, \texttt{FedAdagrad}, and \texttt{FedYogi}, which address data heterogeneity, \texttt{PLFL} \cite{bose2024addressing} for personalized FL, as well as \texttt{IIADMM} \cite{ryu2022appfl}, which focuses on efficient privacy preservation. Additionally, for asynchronous aggregation, \textsc{Appfl} includes strategies such as \texttt{FedAsync}, \texttt{FedBuff}, and \texttt{FedCompass}. This diverse suite of options ensures that \textsc{Appfl} can accommodate a variety of needs and scenarios in FL, illustrating its adaptability and comprehensive approach to FL challenges.

\subsection{Privacy Preservation and Authentication}
To tackle security concerns in FL, \textsc{Appfl} offers solutions that span both algorithmic and system-level measures. Algorithmically, \textsc{Appfl} incorporates differential privacy (DP) algorithms \cite{dwork2006differential} into FL that perturbs the client model parameters with noises before sending to the server, protecting against the reconstruction of confidential training data. A study utilizing \textsc{Appfl} showcases that the usage of DP in FL can effectively mitigate the risk of data reconstruction \cite{hoang2023enabling}.

At the system level, \textsc{Appfl} enhances security through the integration of identity and access management (IAM) services into its communication stack for user authentication and access control for FL experiments. Specifically, Globus Compute itself is already integrated with the Globus authentication service, ensuring that the server dispatches training functions only to clients within a specified Globus group. This setup helps create a secure federation of trusted collaborators, authenticated via institutional emails linked to Globus accounts.

\begin{figure}[h]
\centerline{\includegraphics[width=0.9\linewidth]{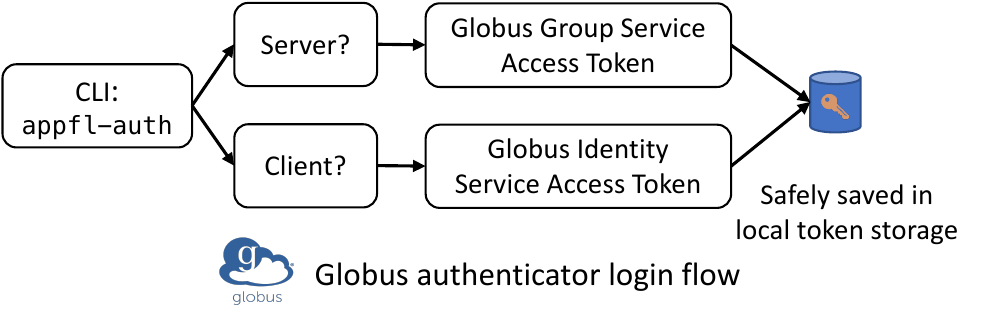}}
\caption{Login flow for the Globus authenticator.}
\label{fig:globus-auth}
\end{figure}

As for gRPC, \textsc{Appfl} utilizes token-based authenticators to verify users. Clients have to attach an access token to each remote procedure call (RPC) request over an SSL-encrypted channel, allowing the server to confirm the user's identity before processing the request. The token-based authenticator consists of two primary functions: one invoked by the client to generate the token prior to sending the RPC request and another invoked by the server to verify the validity of the token upon receipt. This straightforward interface allows developers to effortlessly integrate their own authentication methods tailored to specific use cases and applications. Currently, \textsc{Appfl} supports a Globus authenticator, with its login flow depicted in Figure~\ref{fig:globus-auth}. Users can employ \textsc{Appfl}'s command line interface (CLI), \texttt{appfl-auth}, to perform a one-time login. Depending on the selected role during login, either as an FL server or client, the appropriate Globus access token (Group Service or Identity Service) is requested. The access tokens, along with the corresponding refresh tokens, are securely stored in the client's local token storage. Whenever an FL client makes an RPC request, it attaches its Globus Identity Service token. The FL server uses this token to retrieve the client’s Globus ID and, leveraging its Globus Group Service token, verifies whether the client belongs to the specified Globus group. This robust authentication process ensures a secure and controlled federation for FL experiments, and significantly reduces the risk of malicious attack.

\vspace{-4pt}
\section{Performance Evaluation}
\vspace{-0.5pt}
\begin{figure*}[h]
\centerline{\includegraphics[width=0.9\linewidth]{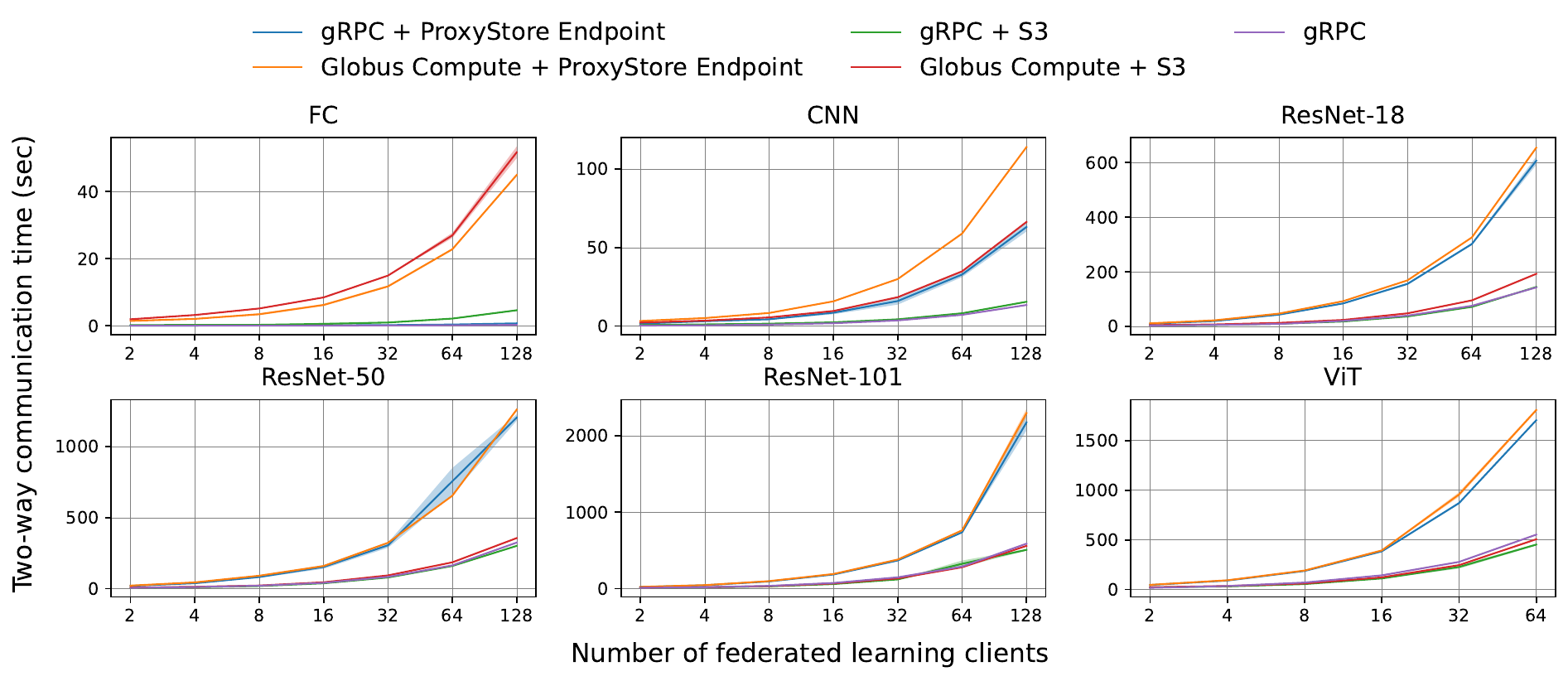}}
\captionsetup{belowskip=-14pt} 
\caption{Efficiency comparison of communication and data transfer protocols: Average two-way communication time per global epoch and the corresponding standard deviation as the number of clients increases exponentially across various models.}
\label{fig:communication-results}
\end{figure*}

In this section, we employ \textsc{Appfl} to benchmark a broad spectrum of components within FL to highlight the advantages of its software design. Specifically, we utilize \textsc{Appfl} to evaluate the communication efficiency of different protocols, data transfer methods, and model compression algorithms. We also explore the impacts of privacy preservation algorithms on the performance of FL-trained models, as well as the training efficiency and resource utilization of various FL strategies with different synchronicity settings. Note that this section does not directly compare \textsc{Appfl} with other frameworks on the capabilities to resolve different FL challenges, as such framework's capability stems from its design, consisting of a suite of components and comprehensive algorithmic and technical solutions to address diverse FL challenges. Table~\ref{tab:frameworks} already summarizes and compares those features among existing frameworks, and these solutions are generally framework-agnostic, provided a framework is capable of supporting them.

\vspace{-4pt}

\subsection{Communication Efficiency}\label{sec:comm-eff}

We evaluate the communication efficiency for different communication and data transfer protocols across different numbers of clients and various model sizes. Table~\ref{tab:modelsize} details the sizes of all models used in our experiments. 

\begin{table}[h]
    \centering
    \footnotesize
    \caption{Sizes of the models used in the experiments.}
    \label{tab:modelsize}
    \begin{tabular}{c c c}
    \toprule
    \textbf{Model} & \textbf{\# Params} & \textbf{Size} \\ \midrule
    1$\times$1 FC & 2 & 8 B\\
    CNN & 1.20M & 4.58 MB \\ 
    ResNet18  & 11.17M & 42.66 MB \\ 
    ResNet50  & 23.52M & 89.93 MB \\
    ResNet101 & 42.51M & 162.58 MB \\
    Vision Transformer & 88.22M & 336.55 MB \\ \bottomrule
    \end{tabular}
\end{table}

In the experiments, each FL client runs on a single core of a CPU node that contains two 64-core AMD EPYC 7763 ``Milan'' CPUs with PCIe Gen4 interfaces and 256 GB of RAM. The node is connected to the NPCF core router and exit infrastructure via two 100 gigabits per second (Gbps) connections. The FL server is hosted on an AWS EC2 {x2iedn.2xlarge} instance, equipped with 8 virtual CPUs, 256 GB of RAM, and up to 25 Gbps connections. We exponentially increase the number of clients from 2 to 128 across all models, except for the ViT model, which  scales only from 2 to 64 because of memory constraints on the client and server hardware. We deploy all FL clients on the same hardware node to avoid performance variability caused by slower or heterogeneous machines, ensuring consistent and reproducible experimental results and accurately isolating the impact of communication protocols. We evaluate gRPC and Globus Compute communication protocols as well as two data transfer methods, AWS S3 buckets and ProxyStore endpoints. Because of the 10 MB data transfer limit with Globus Compute, it is integrated with the other two data transfer protocols rather than being tested in isolation, resulting in five distinct communication pattern combinations. 

Figure~\ref{fig:communication-results} shows the epoch-wise average two-way communication time in seconds for various models using different communication and data transfer protocols. From the plots, we note the following key points: (1) Separating the transmission of data (i.e. model parameters) from task control signals helps communication protocols exceed their maximum data size limitations. (2) Globus Compute consistently incurs longer overheads than gRPC in transmitting control signals, which is a significant factor when the FL model size is small. (3) While data transfer via ProxyStore endpoints generally results in longer communication times, it offers a free and straightforward solution for protocols such as Globus Compute that have message size restrictions. (4) Data transfer through S3 features relatively low latency and also provides a secure, reliable means to store model checkpoints during training, although it incurs some additional costs on AWS.

\vspace{-4pt}

\subsection{Compression Efficiency}
\begin{figure*}[h]
    \centering
    \begin{subfigure}[b]{0.33\linewidth}
        \centering
        \includegraphics[width=\linewidth]{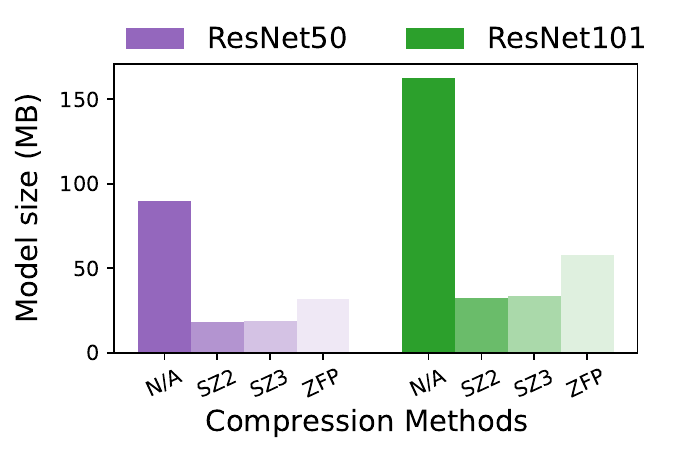}
        \caption{}
        \label{fig:compress-size}
    \end{subfigure}
    \begin{subfigure}[b]{0.6\linewidth}
        \centering
        \includegraphics[width=\linewidth]{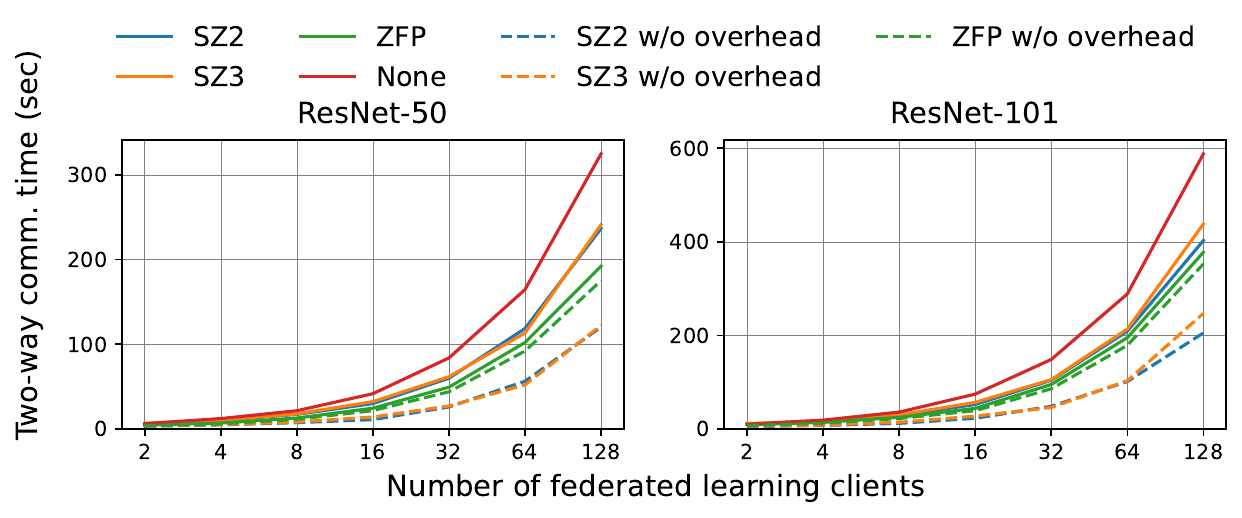}
        \caption{}
        \label{fig:compress-time}
    \end{subfigure}    
    \captionsetup{belowskip=-16pt} 
    \caption{(a) Model sizes for ResNet-50 and ResNet-101 using different lossy compression methods with a relative error bound of 0.01. (b) gRPC two-way communication time for ResNet-50 and ResNet-101 using different lossy compressors, with and without the compression and decompression overhead.}
    \label{fig:compression-results}
\end{figure*}

We assess the efficiency of various data compression algorithms integrated within \textsc{Appfl}. Specifically, we utilize the lossless compressor blosc for tensors with less than 1,024 parameters and lossy compressors SZ2 \cite{liang2018error}, SZ3 \cite{liang2022sz3}, and ZFP \cite{lindstrom2014fixed}, each with a relative error bound of 0.01, for larger tensors. The experiments adhere to the same hardware configurations described in Subsection~\ref{sec:comm-eff}. We conduct experiments on ResNet-50 and ResNet-101 models, scaling client numbers from 2 to 128. Figure~\ref{fig:compress-size} illustrates the reduction in model sizes by 3 to 5 times using different lossy compressors. Notably, previous studies have shown that such levels of lossy compression can preserve model accuracy within a 0.5\% margin of uncompressed results \cite{wilkins2024fedsz}. Figure~\ref{fig:compress-time} presents the two-way communication times via gRPC for the two models using various compressors. Solid lines represent times with compression and decompression overheads, whereas dotted lines depict times without. The comparison reveals significant overhead, particularly with SZ2 and SZ3. Despite this, the use of compressors notably reduces communication costs and overall two-way communication times, even under high-bandwidth conditions for both clients and the server. 

\subsection{Privacy Preservation}
\begin{table}[h]
    \centering
    \scriptsize
    \setlength{\tabcolsep}{1.9pt} 
    \caption{Overview of selected tasks from FLamby.}
    \label{tab:flamby}
    \begin{tabular}{r c c c c}
    \toprule
    & Fed-TCGA-BRCA & Fed-Heart-Disease & Fed-IXI & Fed-ISIC2019 \\
    \midrule
    Input & Patient info & Patient info & T1WI & Dermoscopy \\
    Prediction & Risk of death & Heart disease & Brain mask & Melanoma class \\
    Task type & Regression & Classification & 3D Segmentation & Classification \\
    Model & Cox model & Logistic Reg. & 3D U-Net & EfficientNet \\
    Metric & C-index & Accuracy & DICE  & Balanced Acc. \\
    \# Clients & 6 & 4 & 3 & 6 \\
    \bottomrule
    \end{tabular}
\end{table}

\begin{figure}[h]
    \centerline{\includegraphics[width=
    0.99\linewidth]{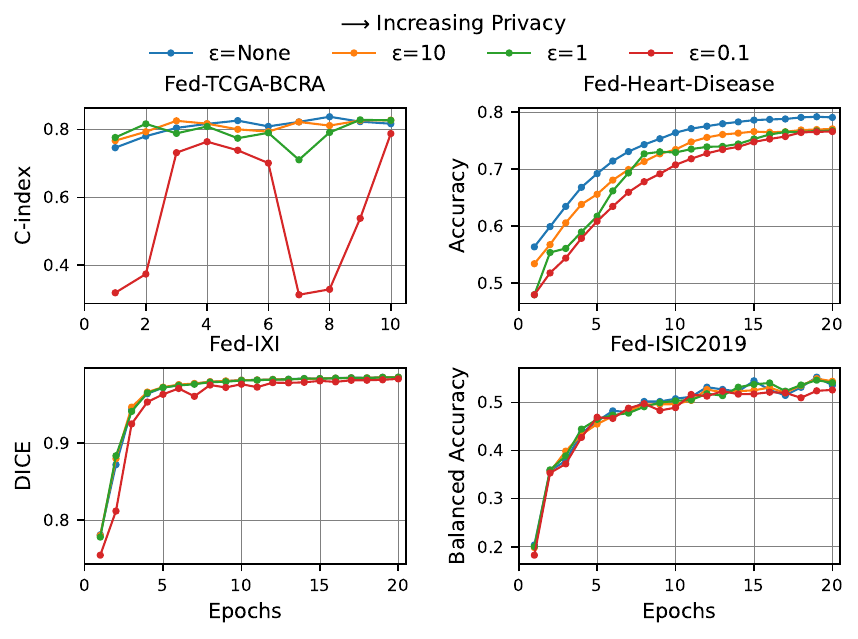}}
    \caption{Change of model performance throughout the FL training on the selected FLamby tasks at different $\epsilon$ values.}
    \label{fig:privacy-results}
\end{figure}
In this subsection we study the impact of differential privacy (DP) techniques on the performance of models trained via FL. We select four tasks in medical domains, where data privacy is paramount, from the FLamby benchmark containing naturally split medical datasets \cite{ogier2022flamby}. Table~\ref{tab:flamby} provides an overview of these tasks. We assess model performance across varying values of privacy loss parameter $\epsilon$, a measure of how much privacy is lost when using DP algorithms, with lower $\epsilon$ values signifying larger added noises and enhanced privacy. Figure~\ref{fig:privacy-results} shows the change of model performance throughout the FL training process for these tasks at different $\epsilon$ values. The performance metrics represent the average outcomes of five independent trials with different random seeds. The results indicate that a decrease in $\epsilon$ values, corresponding to increased privacy preservation, leads to varying degrees of performance degradation across various models and training tasks.

\subsection{Addressing Heterogeneous Clients}
\begin{figure*}[htbp]
    \centering
    \begin{subfigure}[b]{0.3\linewidth}
        \centering
        \includegraphics[width=\linewidth]{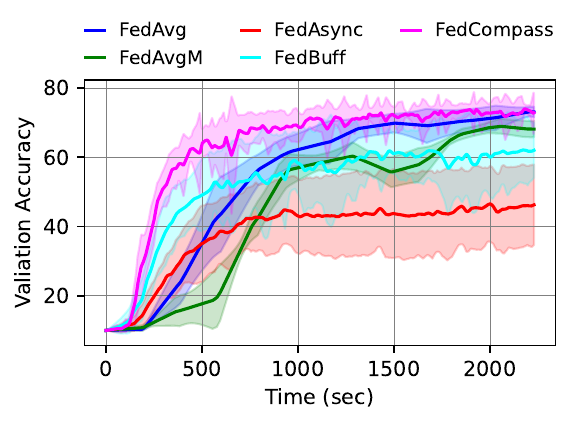}
        \caption{}
        \label{fig:efficiency-acc}
    \end{subfigure}
    \begin{subfigure}[b]{0.3\linewidth}
        \centering
        \includegraphics[width=\linewidth]{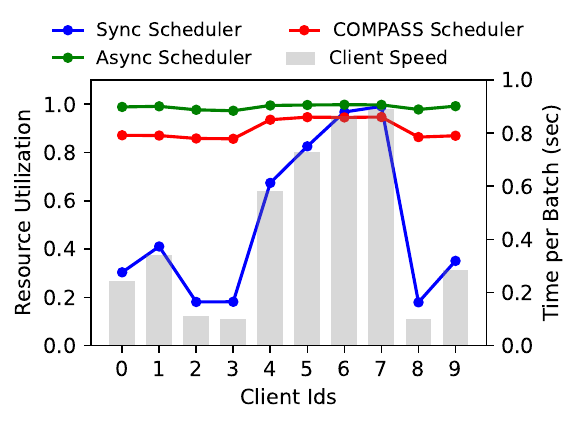}
        \caption{}
        \label{fig:efficiency-util}
    \end{subfigure}
    \begin{subfigure}[b]{0.3\linewidth}
        \centering
        \includegraphics[width=\linewidth]{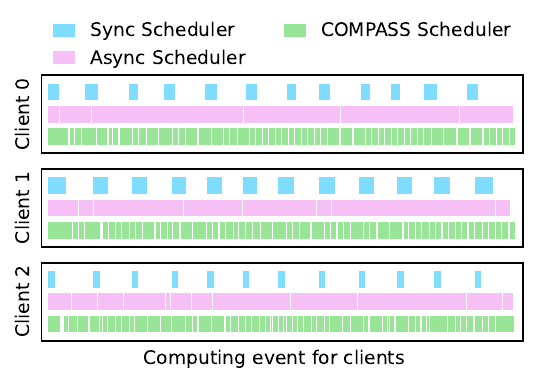}
        \caption{}
        \label{fig:efficiency-event}
    \end{subfigure}
    \captionsetup{belowskip=-15pt} 
    \caption{(a) Average validation accuracy and the corresponding standard deviation on the partitioned CIFAR-10 dataset for different FL algorithms during the training process. (b) Client resource utilization for algorithms using different schedulers and the average training time per batch for different clients. (c) Visualization of computing resource utilization for three clients under different schedulers, where the colored bar represents the computing period and the blank places mean idle times.}
    \label{fig:efficiency}
\end{figure*}
In this subsection we evaluate the performance and efficiency of different FL algorithms under various synchronicity settings. Specifically, we benchmark five FL algorithms: (1) \texttt{FedAvg}, a widely used synchronous algorithm that updates the global model by averaging all client local models; (2) \texttt{FedAvgM}, another synchronous algorithm, which incorporates momentum on top of \texttt{FedAvg}; (3) \texttt{FedAsync}, which asynchronously updates the global model upon receipt of any local model; (4) \texttt{FedBuff}, which is similar to \texttt{FedAsync} but buffers multiple local models before updating the global model; and (5) \texttt{FedCompass}, which introduces a \textit{COMputing Power AwarenesS Scheduler} (\texttt{Compass}) that dynamically adjusts the number of client local training steps based on real-time estimates of client computing power to synchronize the training completion for groups of clients. As for the datasets, we partition the CIFAR-10 dataset, one of the most commonly used datasets in evaluating FL algorithms, into ten client splits in a non-IID manner, with each client holding data from five to seven classes out of ten classes. All clients use Nvidia A100 GPUs for training, and we simulate a group of heterogeneous clients by assigning different average batch processing times from an exponential distribution.

Figure~\ref{fig:efficiency-acc} presents the average validation accuracy and the corresponding standard deviation across five independent runs for each FL algorithm during  training. Key observations from the figure include the following. (1) Asynchronous FL algorithms like \texttt{FedAsync} and \texttt{FedBuff}, which use the vanilla asynchronous scheduler, converge to significantly lower global model accuracy compared with synchronous methods, primarily due to the drifting toward faster clients, as the global model gets more updates from faster clients and slower clients' models become stale. (2) Synchronous algorithms exhibit slower convergence as the server has to wait for the slow clients for aggregation. (3) \texttt{FedCompass} effectively addresses substantial client drift issues and attains high global model accuracy by ensuring nearly simultaneous model arrivals for grouped aggregation. It also achieves quicker convergence than synchronous methods without extensive waiting. To our best knowledge, no existing FL frameworks seamlessly support advanced FL scheduling algorithms such as \texttt{FedCompass} without architectural modifications, underscoring the extensibility of the \textsc{Appfl} framework.

Figure~\ref{fig:efficiency-util} shows the average training time per batch for the ten clients involved in the FL training, as well as the resource utilization, calculated as the ratio of client compute time to total training time, for algorithms using the synchronous, vanilla asynchronous, and \texttt{Compass} asynchronous scheduler. The synchronous scheduler shows the lowest client resource utilization, correlating with training time per batch: the quicker the client, the lower the utilization. In contrast, the vanilla asynchronous scheduler, which immediately sends any received local model for aggregation and returns the updated global model, allows client resource utilization to approach 100\%. Despite full utilization, however, this method results in poorly performing models due to client drift. The \texttt{Compass} scheduler, by estimating client speeds and adjusting training steps accordingly, maintains approximately 90\% resource utilization and reduces client drift through timely grouped aggregations. Figure~\ref{fig:efficiency-event} visualizes the resource utilization for three clients under different scheduling scenarios, highlighting the significant resource underutilization of the synchronous scheduler compared with the asynchronous alternatives when client computing resources vary widely. 

\section{Case Study: Extensibility Demonstration}
To highlight the versatility and extensibility of the \textsc{Appfl} framework across various FL applications, we present case studies on three distinct FL variants: vertical FL, hierarchical FL, and decentralized FL, all built upon the \textsc{Appfl} framework, illustrating how it can be adapted to different FL paradigms.

\subsection{Vertical Federated Learning}
\begin{figure}[h]
    \centering
    \begin{subfigure}[b]{0.49\linewidth}
        \centering
        \includegraphics[width=\linewidth]{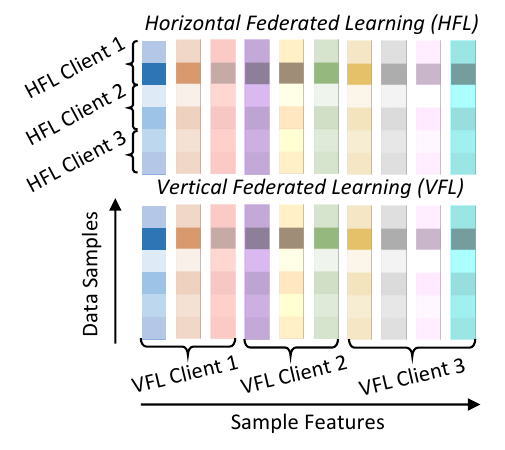}
        \caption{}
        \label{fig:vfl-data}
    \end{subfigure}
    \hfill
    \begin{subfigure}[b]{0.49\linewidth}
        \centering
        \includegraphics[width=\linewidth]{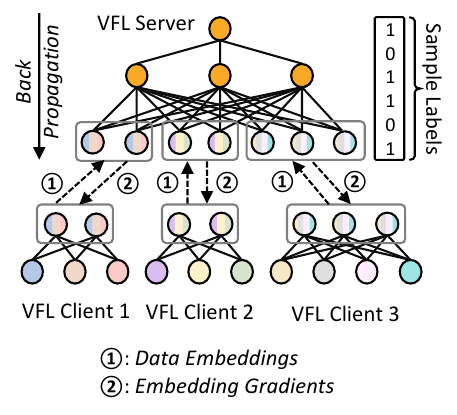}
        \caption{}
        \label{fig:vfl-process}
    \end{subfigure}
    \caption{(a) Comparison of client training data distribution in HFL and VFL. (b) Overview of the VFL process. }
    \label{fig:vfl}
\end{figure}

\begin{figure}[t]
    \centering
    \begin{subfigure}[c]{0.35\linewidth}
        \centering
        \includegraphics[width=\linewidth]{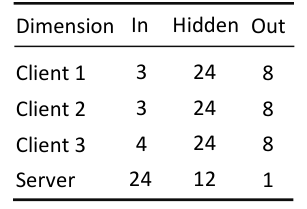}
        \caption{}
        \label{fig:vfl-model}
    \end{subfigure}
    \hfill
    \begin{subfigure}[c]{0.63\linewidth}
        \centering
        \includegraphics[width=\linewidth]{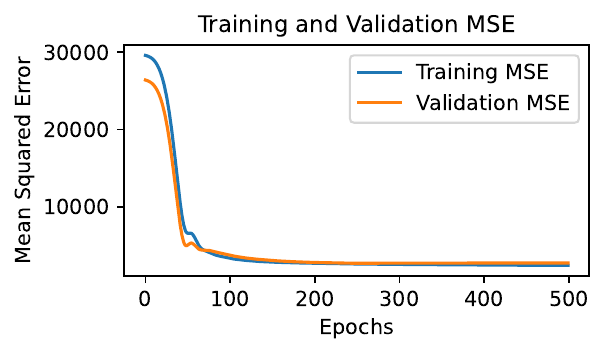}
        \caption{}
        \label{fig:vfl-result}
    \end{subfigure}
    \caption{(a) Input, hidden, and output dimensions of two-layer perceptrons for the VFL clients and server. (b) Training and validation MSE during the VFL training process.}
    \label{fig:vfl2}
\end{figure}
Vertical federated learning (VFL) is a specialized paradigm of FL where different clients hold distinct features from the same dataset \cite{liu2024vertical}. Unlike traditional FL (i.e., horizontal FL) dealing with the same feature space across diverse data samples, VFL enables collaboration among clients that have partially overlapping or non-overlapping features but share the same sample IDs, as illustrated in Figure~\ref{fig:vfl-data}. Figure~\ref{fig:vfl-process} depicts a typical VFL process. In VFL, rather than training the same model architecture, each client possesses its embedding model to process its local data sample features and then sends their embeddings to the server. The server, holding the labels of the client data samples, concatenates the received embeddings to train a central model. It then sends the gradients of the feature embeddings back to the corresponding clients, enabling them to update their local embedding models accordingly. 

\textsc{Appfl} seamlessly supports VFL by providing the VFL trainer and aggregator in the corresponding modules. In this case study, we use the diabetes datasets from the \texttt{scikit-learn} library, which contains ten features of 442 data samples. The labels, ranging from 25 to 346, are the responses of interest that quantitatively measure the disease progression. We split the dataset into 80\% for training and 20\% for validation and use three VFL clients, where clients 1 and 2 possess three patient features and client 3 possesses four. Each of the three clients as well as the server employs a two-layer perceptron with ReLU nonlinear activation  as their embedding models. Figure~\ref{fig:vfl-model} presents the input, hidden, and output dimensions of these models. During the training, the server model is updated based on the mean squared error (MSE) loss between the labels and predictions, using the Adam optimizer with a learning rate of 0.01. Figure~\ref{fig:vfl-result} shows the training and validation MSE throughout the training.

\subsection{Hierarchical Federated Learning}

Hierarchical federated learning (HierFL) is also a special type of FL that introduces an additional role, the intermediate server (edge server). This server first aggregates local model parameters from connected clients or child intermediate servers and then forwards the aggregated model to the parent server for further aggregation \cite{abad2020hierarchical}. 
HierFL is particularly beneficial when FL clients are geographically clustered, since placing an intermediate server for these clusters can significantly improve overall communication efficiency. 
To support HierFL in \textsc{Appfl}, in addition to the general server agent for the root server and the client agent for the clients, we define an intermediate server agent, inherited from the server agent, which handles FL-related requests from connected clients or child intermediate servers by interacting with its parent server. 

\begin{figure}[h]
    \centering
    \begin{subfigure}[b]{0.37\linewidth}
        \centering
        \includegraphics[width=\linewidth]{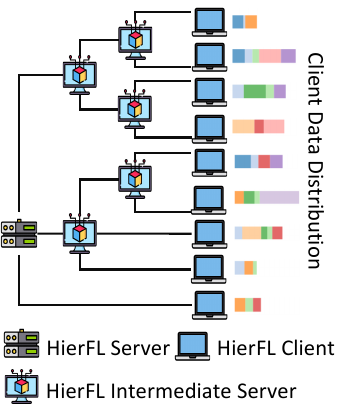}
        \caption{}
        \label{fig:hfl-topo}
    \end{subfigure}
    \hfill
    \begin{subfigure}[b]{0.59\linewidth}
        \centering
        \includegraphics[width=\linewidth]{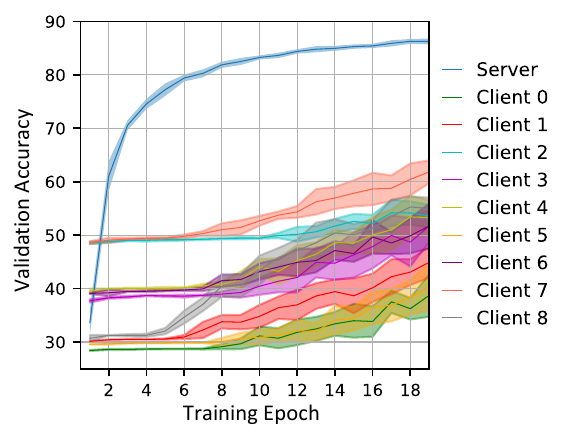}
        \caption{}
        \label{fig:hfl-result}
    \end{subfigure}
    \caption{(a) Topology of the multi-layer HierFL experiments. (b) HierFL validation accuracy for the server and client models, where the accuracy of client models is evaluated after each local training round. }
    \label{fig:hfl}
\end{figure}

This case study conducts four-tier HierFL experiments involving nine clients, five intermediate servers, and one root server. The MNIST dataset is partitioned into nine heterogeneous splits, with each client containing training data for only 3-5 classes. Figure~\ref{fig:hfl-topo} illustrates the topology of the experiment and the client data distribution. Training is conducted over 20 global epochs, with each client performing 100 local steps per epoch using a batch size of 64 and the Adam optimizer with a learning rate of 0.001. The experiments are repeated five times. Figure~\ref{fig:hfl-result} presents the average validation accuracy and standard deviation for both the server model and each client’s local model on the MNIST validation set. We note that the client models are evaluated after local training. Since each client has data for only three to five classes, their local models perform significantly worse than the global model, highlighting the advantages of federated learning in leveraging data from distributed clients to train a more robust ML model.

\vspace{-3pt}

\subsection{Decentralized Federated Learning}

\begin{figure}[h]
    \centering
    \begin{subfigure}[b]{0.335\linewidth}
        \centering
        \includegraphics[width=\linewidth]{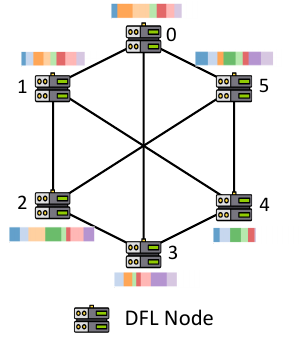}
        \caption{}
        \label{fig:dfl-topo}
    \end{subfigure}
    \hfill
    \begin{subfigure}[b]{0.575\linewidth}
        \centering
        \includegraphics[width=\linewidth]{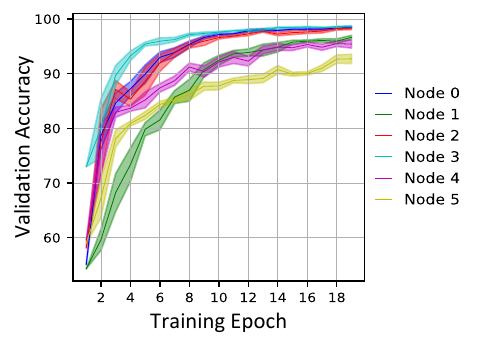}
        \caption{}
        \label{fig:dfl-result}
    \end{subfigure}
    \caption{(a) Topology of the DFL experiments. (b) DFL validation accuracy for the DFL nodes, evaluated after aggregating the local models of the neighbor DFL nodes.}
    \label{fig:dfl}
\end{figure}

Decentralized federated learning (DFL) is another FL variant that eliminates the need for a central server. Instead, each node trains its local model, requests model parameters from neighboring clients, and aggregates these with its local model \cite{lalitha2018fully}. \textsc{Appfl} supports DFL by implementing a DFL node agent that inherits functionalities of both an FL client and server, enabling it to train local models and handle requests from neighboring clients. This case study sets up DFL experiments with six nodes, where each node has three neighbors, as shown in Figure~\ref{fig:dfl-topo}. Each node holds a heterogeneously partitioned MNIST dataset with six to eight classes and trains the model for 20 epochs. During each epoch, the node updates its model for 100 steps with a batch size of 64 using the Adam optimizer with a learning rate of 0.001, then aggregates its local model with those of its three neighbors. The experiment is repeated five times.  Figure~\ref{fig:dfl-result} presents the average validation accuracy and its standard deviation on the MNIST validation set across the training process for the six DFL nodes.

\vspace{-8pt}

\section{Conclusion and Future Work}
In this paper, we present the recent advancements in \textsc{Appfl}, a federated learning framework to simplify FL usage by offering comprehensive solutions to various challenges and to advance FL research through an easy-to-use, modular interface that facilitates the seamless integration of new algorithms. We demonstrate the capability and extensibility of \textsc{Appfl} by employing it to benchmark various FL components and provide case studies across different FL variants. \textsc{Appfl} is open-sourced under the MIT License, and we actively encourage contributions from the community. In our future work, we plan to incorporate more advanced privacy-enhancing technologies into the framework, such as secure multi-party computation, homomorphic encryption, and trusted execution environments, to further ensure the security of FL experiments. We also aim to employ \textsc{Appfl} for training larger-scale foundation models by leveraging private data from multiple data silos.
\section*{Acknowledgment}
This material is based upon work supported by the U.S. Department of Energy, Office of Science, under contract number DE-AC02-06CH11357. This research utilizes computing resources provided by the National Artificial Intelligence Research Resource (NAIRR) Pilot, supported by award NAIRR240008. We also gratefully acknowledge Amazon Web Services (AWS) for providing cloud computing credits that were used to assist with benchmarking efforts for this paper.
\bibliographystyle{IEEEtran}
\bibliography{IEEEabrv,mybibfile}

\end{document}